\documentclass[sn-aps]{sn-jnl}

\usepackage{graphicx}%
\usepackage{multirow}%
\usepackage{amsmath,amssymb,amsfonts}%
\usepackage{amsthm}%
\usepackage{mathrsfs}%
\usepackage[title]{appendix}%
\usepackage{xcolor}%
\usepackage{textcomp}%
\usepackage{manyfoot}%
\usepackage{booktabs}%
\usepackage{algorithm}%
\usepackage{algorithmicx}%
\usepackage{algpseudocode}%
\usepackage{listings}%

\newcommand\bm{\boldsymbol}
\newcommand\bnabla{\boldsymbol{\nabla}}
\newcommand\bcdot{\boldsymbol{\cdot}}
\newcommand{\grad}{\bnabla} 
\newcommand{\curl}{\bnabla\times} 

\newcommand{\revision}[1]{#1}

\begin{document}

\title{A critical assessment of reinforcement learning methods for microswimmer navigation in complex flows}

\author[1]{\fnm{Selim} \sur{Mecanna}}\email{selim.mecanna@centrale-med.fr}
\author*[1]{\fnm{Aurore} \sur{Loisy}}\email{aurore.loisy@irphe.univ-mrs.fr}
\author*[1]{\fnm{Christophe} \sur{Eloy}}\email{christophe.eloy@centrale-med.fr}
\affil[1]{Aix Marseille Univ, CNRS, Centrale Med, IRPHE, Marseille, France}

\abstract{
Navigating in a fluid flow while being carried by it, using only information accessible from on-board sensors, is a problem commonly faced by small planktonic organisms. It is also directly relevant to autonomous robots deployed in the oceans. 
In the last ten years, the fluid mechanics community has widely adopted reinforcement learning, often in the form of its simplest implementations, to address this challenge. But it is unclear how good are the strategies learned by these algorithms. 
In this paper, we perform a quantitative assessment of reinforcement learning methods applied to navigation in partially observable flows. 
We first introduce a well-posed problem of directional navigation for which a quasi-optimal policy is known analytically.
We then report on the poor performance and robustness of commonly used algorithms (Q-Learning, Advantage Actor Critic) in flows regularly encountered in the literature: Taylor-Green vortices, Arnold-Beltrami-Childress flow, and two-dimensional turbulence.
We show that they are vastly surpassed by PPO (Proximal Policy Optimization), a more advanced algorithm that has established dominance across a wide range of benchmarks in the reinforcement learning community.
In particular, our custom implementation of PPO matches the theoretical quasi-optimal performance in turbulent flow and does so in a robust manner. Reaching this result required the use of several additional techniques, such as vectorized environments and generalized advantage estimation, as well as hyperparameter optimization. 
This study demonstrates the importance of algorithm selection, implementation details, and fine-tuning for discovering truly smart autonomous navigation strategies in complex flows.
}


\maketitle

\section{Introduction}




The development of artificial microswimmers with navigation capabilities has been an intense topic of research in recent years \citep{MS1,MS2,Muinos-Landin2021}. When such robots with limited self-propulsion abilities are carried by a fluid flow, navigation becomes notoriously harder. This is the kind of challenge faced by robots deployed in the oceans for environmental monitoring purposes. Ideally, these drifting robots would be able to exploit background currents to travel more efficiently while relying only on data from their on-board sensors. The very same problem is also faced by plankton: these small organisms that drift with currents may be able to exploit hydrodynamic cues to migrate efficiently over long distances \citep{BioGrads1,BioGrads2,Monthiller2022}.

If the agent had global information about the flow, optimal control theory could be used to find optimal trajectories (a problem known as Zermelo's navigation problem \citep{Zermelo}). But when the agent can only sense the flow \emph{locally} (that is, has only access to a \emph{partial observation}), optimal control theory can no longer be used. 
This problem becomes a model-free partially observable Markov decision process (POMDP). 
Such problems are usually well-suited for reinforcement learning, a data-driven alternative to optimal control that allows an agent to be trained at solving a task through repeated interactions with its environment.

In the last ten years, navigation in partially observable flows has attracted considerable attention in the fluid mechanics community \citep{Gazzola2014,Reddy2016a,Gazzola2016,Colabrese2017,Gustavsson2017,Reddy2018,Colabrese2018,Verma2018,Alageshan2020,Gunnarson2021,Mandralis2021,Qiu2022a,Qiu2022b,Calascibetta2023a,ElKhiyati2023,Sankaewtong2023,Gunnarson2024,Mousavi2025,Jiao2025}, who adopted reinforcement learning techniques to develop "smart" navigation strategies.
A variety of problems have been addressed, often inspired by biology. They include exploiting the flow to travel more efficiently \citep{Reddy2016a,Colabrese2017,Gustavsson2017,Reddy2018,Qiu2022a}, maintaining stable collective formations \citep{Gazzola2014,Gazzola2016,Verma2018}, catching a passive target \citep{Alageshan2020}, reducing chaotic dispersion \citep{Calascibetta2023a}, or targeting specific regions of the flow \citep{Colabrese2018,Gunnarson2024,Mousavi2025}. In parallel, various physical models of the agent have been used, ranging from simple self-propelled point particles to deformable bodies with fluid-structure interactions (e.g., \citep{Verma2018,ElKhiyati2023}).

\revision{A significant part} of recent studies still use tabular Q-Learning, a classic learning algorithm \citep{QL} that has long been superseded by "deep" methods in the reinforcement learning community.
A recent paper highlighted the limitations of such `vanilla'\footnote{`vanilla' refers to the most basic, textbook implementation of an algorithm, without the use of any extra technique to improve on its performance} learning algorithms for discovering good strategies in complex flows, by showing that none of them could match the performance of a simple heuristic strategy obtained from physical intuition \citep{ElKhiyati2023}. 
Therefore, the quality (with respect to optimality) of the learned strategies obtained so far in various navigation problems is uncertain. Previous work has shown that vanilla algorithms can find a solution to various navigation problems. But what does it take to find a \emph{good} solution (close to optimality)? 

In this paper, we provide a rigorous assessment of reinforcement learning as a tool to discover navigation strategies for microswimmers in complex flows. We compare three algorithms: two are representative of those used in prior work on smart microswimmers (Q-Learning \citep{QL} and Advantage Actor-Critic \citep{A2C}), the last one is one the best modern algorithms for reinforcement learning and has demonstrated its capabilities across a wide range of domains (Proximal Policy Optimization \citep{PPO}). We benchmark these algorithms on a simple navigation problem in three different flows that are representative of those used in prior work: Taylor-Green vortices, Arnold-Beltrami-Childress flow, and two-dimensional unsteady turbulence.

Our study reveals that Q-Learning and A2C (Advantage Actor Critic) algorithms, despite being still routinely used for this purpose, actually perform rather poorly on navigation in partially observable flows. In contrast, we show that a custom implementation of PPO (Proximal Policy Optimization ) allows learning a policy that matches the near-optimal performance. This work demonstrates that deep reinforcement learning is indeed a promising path toward autonomous navigation in flows, but only at the price of careful algorithm selection, implementation, and tuning.

The paper is organized as follows. We start in Section~\ref{sec:problem} by defining a well-posed benchmark navigation problem for which a near-optimal policy is known analytically, and introduce the three flow environments. In Section~\ref{sec:rl_algo}, we present the reinforcement learning algorithms used in this paper. In Section~\ref{sec:results}, we report on the performance and robustness of these algorithms in the three different flows. We conclude with a summary and discussion in Section~\ref{sec:conc}.

\section{Navigation in complex flows}
\label{sec:problem}

\subsection{Statement of the problem}

We consider a swimming agent trying to go as far as possible in the upward $\hat{\bm{z}}$ direction. This task is representative of the diel vertical migration of plankton, and more generally of long-distance navigation to a target point (here moved to infinity) without the additional difficulties associated with sparse rewards. 

The agent is modeled as an inertialess point-like particle swimming at a constant swimming speed $v$ while being advected by the surrounding flow $\bm{u}(\bm{x}, t)$. \revision{For simplicity, we assume that the agent does not modify the background flow (one-way coupling), an approximation valid in the dilute limit as the agent perturbation to the flow is decreasing as an inverse power law. The agent} can only control its swimming direction $\hat{\bm{p}}(t)$, a unit vector, every $\Delta t$ (decision time). The readjustment of its swimming direction is instantaneous (no reorientation delay). Under these assumptions, the agent motion is governed by the following equation:
\begin{equation} \label{eq:dynamics}
\bm{X}(t_{n+1}) = \bm{X}(t_{n}) + \int_{t_n}^{t_{n+1}}\left(\bm{u}(\bm{X}, t) + v \hat{\bm{p}}(t_{n})\right)d t, \quad \bm{X}(t_0) = \bm{X}_0
\end{equation}
where $\bm{X}(t)$ is the position of the agent at time $t$, $\bm{u}(\bm{x},t)$ is the flow velocity field (incompressible and with zero mean flow), and $\Delta t = t^{n+1} - t^n$.
The agent initial position $\bm{X}_0$ is randomly initialized in the flow, and the starting time $t_0$ is also chosen randomly (when the flow is unsteady). The agent's swimming speed $v$ is chosen as roughly half the typical flow speed (cf. Section \ref{sec:env_def}). This choice ensures that the agent significantly drifts with the flow, while keeping learning computationally inexpensive for the purpose of this systematic benchmark.

In order to choose its swimming direction $\hat{\bm{p}}(t)$ at best, the agent has access to local flow information $\bm{G}(\bm{X}(t), t)$. This observable is chosen to be the local velocity gradient tensor $\grad \bm{u}(\bm{X}(t), t)$ (or a related quantity, depending on the flow considered, cf. Section \ref{sec:env_def}). Indeed only flow \emph{gradients}, rather than the flow itself, can be measured by an agent drifting with the flow. 

The goal of the agent is to maximize the total distance traveled along the target direction $\hat{\bm{z}}$ over an episode of duration $T=t_{f}-t_{0}$, and averaged over all random initial conditions.
We denote this metric $Z$, formally defined as
\begin{equation} \label{eq:goal}
   Z = <(\bm{X}(t_f) - \bm{X}_0) \cdot \hat{\bm{z}}>
\end{equation}
where the brackets indicate the average.
To summarize, we are looking for the control (called policy in reinforcement learning) $\hat{\bm{p}}(\bm{G})$ that maximizes the objective function $Z$ under the dynamics given by Eq.~(\ref{eq:dynamics}).

This problem is simple enough to have a known analytical approximate solution (cf. next section) while retaining the complexity inherent to plankton-like navigation. For these reasons, it provides a well-posed benchmark problem for evaluating the capabilities of reinforcement learning applied to autonomous navigation in flows.

\subsection{Analytical baselines} 
\label{sec:analytical_baselines}

Two heuristic policies are considered as baselines: the naive policy and the surfing policy.

The \emph{naive} policy consists in always swimming upward:
\begin{equation} 
\hat{\bm{p}}=\hat{\bm{z}},
\end{equation}
resulting in average travelled distance $Z=vT$.
This naive policy is a weak baseline, representative of baselines used in prior work applying vanilla reinforcement learning algorithms for navigation in flows. 

The \emph{surfing} policy is a recently proposed policy that has been shown to significantly improve upon the naive policy \citep{Monthiller2022}. It reads
\begin{equation} \label{eq:surf}
    \hat{\bm{p}} = \bm{\lambda} / \lVert  \bm{\lambda} \rVert, \qquad
    \bm{\lambda} = \left[\exp\left(\tau^*\bm{G}\right)\right]\cdot\hat{\bm{z}}
\end{equation}
where $\exp$ is the matrix exponential. The parameter $\tau^*$ has a physical meaning: it quantifies the mean correlation time of $\bm{G}$ as observed by the agent along its trajectory. For all practical purposes $\tau^*$ can be seen as a free parameter of the surfing policy that can be manually optimized for each flow (cf. Fig.~\ref{fig:surf_opt}).
The surfing policy provides a strong baseline that reinforcement learning should at least match in order to be considered, in our view, as a suitable method for autonomous navigation in flows. Its name comes from its physical interpretation \citep{Monthiller2022}, as the agent `surfs' on beneficial upward currents.

These two heuristics are, respectively, zeroth- and first-order analytical approximations of the optimal control for this problem, as obtained from Pontryagin's maximum principle. Indeed, our optimization problem can be reformulated as an ordinary differential equation for the adjoint $\bm{\lambda}$:
\begin{align}
\label{eq:adjoint}
 \frac{d\bm{\lambda}(t)}{dt} = - \grad \bm{u}(\bm{X}(t), t) \bcdot \bm{\lambda}(t), \quad \bm{\lambda}(t_f) = \hat{\bm{z}}
\end{align}
which solution is
\begin{equation}\label{eq:lambda_t}
 \bm{\lambda}(t) = \left[ \exp \left( \int_0^{t_f-t}  \grad \bm{u} (\bm{X}(t + \tau), t + \tau) d \tau \right) \right] \bcdot \hat{\bm{z}}
\end{equation}
where $\bm{X}(t)$ is the solution of Eq.~(\ref{eq:dynamics}). In fluid flows, $\grad \bm{u}$ is generally time-correlated over a finite time $\tau^*$. This allows us to approximate the integral in Eq.~(\ref{eq:lambda_t}) by $\tau^* \grad \bm{u}(\bm{X},t)$. The surfing policy, given by Eq.~(\ref{eq:surf}), immediately follows after replacing $\grad \bm{u}$ by $\bm{G}$. Note that neglecting the existence of correlations in the flow amounts to setting $\tau^*=0$, which gives the naive policy.

\subsection{Flows}

We consider three different carrier flows, which are canonical flows commonly used in fluid mechanics: Taylor-Green vortices (TGV), Arnold-Beltrami-Childress flow (ABC), and two-dimensional unsteady turbulence (TURB). These flows provide training environments of increasing difficulty and realism, as they exhibit an increasing number of the key features of real flows: coherent structures (TGV, ABC, TURB), chaotic dynamics (ABC, TURB), and unsteadiness (TURB). In the following we define these flows using standard Cartesian coordinates (x,y,z), with z the coordinate of the target direction.

The TGV flow, illustrated in Fig.~\ref{fig:TGV} (top left), consists of a lattice of counter-rotating vortices. It is an analytical steady solution to the 2D Navier-Stokes equations. It reads:
\begin{align*}
    u_x & = -U \cos(x)\sin(z),\\
    u_z & = U \sin(x)\cos(z)
\end{align*}
where we set $U=0.5$. This flow has been used in, e.g., Refs \cite{Colabrese2017,Qiu2022a,ElKhiyati2023}.

The ABC flow, illustrated in Fig.~\ref{fig:ABC} (top left), is a 3D steady flow characterized by coherent tube-like structures separated by a chaotic region. It is a steady solution to the three-dimensional Euler equations (a particular case of the Navier-Stokes equations with zero viscosity). It reads
\begin{align*}
    u_x & = A\sin(z) + C\cos(y),\\
    u_y & = B\sin(x) + A\cos(z),\\
    u_z & = C\sin(y) + B\cos(x),
\end{align*}
where we set $A=\sqrt{3}$, $B=\sqrt{2}$ and $C=1$. Similar ABC flows have been used in Refs \cite{Gustavsson2017,Colabrese2018}.

 The TURB flow, illustrated in Fig.~\ref{fig:TURB} (top left), is an unsteady, statistically stationary two-dimensional turbulent flow obtained by numerical simulation of the Navier-Stokes equations \citep{Boffetta2012}. This multiscale chaotic flow features moving vortical structures that have a finite lifetime: they unpredictably appear, evolve and vanish. We simulated the flow evolution in the direct cascade regime using a standard pseudo-spectral solver on $256^2$ collocation points and a large scale stochastic forcing. The characteristic flow velocity is $u_{\mathrm{rms}}=3.78$ and the characteristic time scale of the flow (eddy turn-over time) is $\tau_\omega = \omega_{\mathrm{rms}}^{-1} = 0.11$ with $\bm{\omega}=\curl \bm{u}$ the vorticity. Similar 2D turbulent flows have been used in Refs \cite{Alageshan2020,Biferale2019}. 
 
All these flows are $2\pi$-periodic in all directions: when the agent is at position $\bm{X}(t)$, the flow at this location is given by $\bm{u}(\bm{X}(t) \mod 2 \pi, t)$.

\subsection{Environment parameters}
\label{sec:env_def}

The three flows, together with the agent swimming speed $v$, the size of the time step $\Delta t$, the episode duration $T$, and the observable $\bm{G}$, define our three environments. The parameters used are summarized in Table \ref{tab:env_params}.

In TGV, we set $v=u_{\mathrm{max}}/2$ and an episode consists of $4000$ time steps. The observable is $\bm{G} = \grad \bm{u}$. Due to symmetries, only two components of the velocity gradient are independent: these two components form the observation given to the agent, making the observation space two-dimensional.

In ABC, we set $v=u_{\mathrm{max}}/2$ and an episode consists of $2000$ time steps. The observable is the anti-symmetric part of the velocity gradient: $\bm{G} = \frac{1}{2}\left(\grad \bm{u} - \grad \bm{u}^{T}\right)$, which three independent components are proportional to the components of vorticity $\bm{\omega} = \nabla \times \bm{u}$. The observation space is therefore three-dimensional. This choice of observable is motivated by consistency with prior work on ABC flow where vorticity was chosen \citep{Gustavsson2017}. Note that in ABC flow, $\bm{\omega}$ and $\bm{u}$ are equal up to a constant. \revision{The same observable $\bm{G}$, rather than full velocity gradient tensor, is also used for the surfing policy in this environment.}

In TURB, the agent speed is set to $v \approx u_{\mathrm{rms}}/2$, and an episode consists of 500 time steps \revision{(the typical time scale of the flow $\tau_\omega$ is roughly 11 time steps). Turbulent simulation data has been generated for a total duration of 5000 time steps, split into 4000 for training and 1000 for testing. The initial time step that defines the start of an episode is chosen randomly in [0, 3500] for training and [4000, 4500] for testing.} The observable is $\bm{G} = \grad \bm{u}$. Due to flow incompressibility, only three components are independent, making the observation space three-dimensional.

Note that reducing the observation space to independent components is not key: using all the components of $\bm{G}$ yields identical results to those presented hereinafter.

\begin{table}
\begin{tabular}{@{}lrrrrrr@{}}
\toprule
environment & $v$  & $u$ & $\Delta t$ & $T$ & $\bm{G}$ & $\tau^*$ \\
\midrule
TGV & $0.25$ & $0.50$       & $0.01$ & $40.0$  & $\{\partial_x u_x, \partial_x u_z\}$ & $2.0$ \\
ABC & $1.5$ & $3.0$       & $0.01$ & $20.0$  & $\{\omega_x, \omega_y, \omega_z\}$ & $0.72$ \\
TURB & $2.0$ & $3.78$       & $0.01$ & $5.0$  & $\{\partial_x u_x, \partial_x u_z, \partial_z u_x\}$ & $0.23$ \\
\botrule
\end{tabular}
\caption{Parameters of the three environments: agent speed $v$, characteristic flow velocity $u$ (max value for TGV and ABC, root-mean-square value for TURB), decision time step $\Delta t$, duration of an episode $T$, observable $\bm{G}$, optimal value of the parameter $\tau^*$ of the surfing policy.}
\label{tab:env_params}%
\end{table}

\section{Reinforcement learning methods}
\label{sec:rl_algo}

In the language of reinforcement learning and related domains, our navigation problem is a partially observable Markov decision process (POMDP). The agent has only access to an observation $o$ ($\bm{G}$ at its position) of the underlying state $s$ of the environment (the entire flow and the agent's position). The action $a$ is the agent's swimming direction $\hat{\bm{p}}$, and the reward $r$ is the distance traveled in the target direction between two successive time steps. Table \ref{tab:pomdpformulation} maps standard POMDP variables to the corresponding navigation variables. As is usual in reinforcement learning, we apply learning algorithms designed for MDP to our POMDP, assimilating the state to the observable, although there is no guarantee anymore that these algorithms will converge to the optimum policy. In the following, we introduce the three algorithms considered in this work: Q-Learning, A2C (Advantage Actor Critic), and PPO (Proximal Policy Optimization).

\begin{table}
\caption{POMDP framework applied to autonomous navigation.}
\label{tab:pomdpformulation}
\begin{tabular}{lr}
\toprule
POMDP variable &  navigation variable \\
\midrule
$s_n$    &  $\{\bm{X}(t_n), \bm{u}(\bm{X},t) \, \forall t\}$ \\
$a_n$ & $\hat{\bm{p}}(t_n)$ \\
$o_n$    & $\bm{G} (\bm{X}(t_n), t_n)$ \\
$r_n$ & $[\bm{X}(t_{n+1}) - \bm{X}(t_n)]  \cdot \hat{\bm{z}}$ \\
\bottomrule
\end{tabular}
\end{table}

Q-Learning is a value-based method, where the state-action value function (or `Q-function') is estimated and the policy is derived directly from it. It is an off-policy algorithm: the policy used to sample the environment is different from the learned policy (in practice, an $\epsilon$-greedy version of the learned policy is used for sampling). In classical Q-Learning, the Q-function is a table, meaning that observations and actions must be discrete. To use this algorithm, we discretize every component of the observation vector $o$ by categorizing each of them into three possible bins such that one third of the data sampled from each environment belongs to each bin.
The actions are discretized into the four (six) Cartesian directions in 2D (3D), that is, $\pm \hat{\bm{z}}$ and the two (four) orthogonal directions.
\revision{Using finer or coarser discretizations may affect the results, but our Q-Learning experiments are only intended to reproduce prior work, and similar discretizations were used in \cite{Colabrese2017, Gustavsson2017, Alageshan2020, Qiu2022a, Qiu2022b, Calascibetta2023a}.}
We use an optimistic initialization of the Q-matrix to enhance exploration, this significantly improved the results compared to an initialization with zeros.

A2C is an actor-critic method: it combines policy-based methods (actor) and value-based methods (critic). Its name stems from the fact that the critic estimates the advantage function, rather than the state-action value function. It is an on-policy algorithm: the learned policy is used to sample the environment. The actor and critic are feedforward neural networks (see Table \ref{tab:nn}), which enable us to use continuous observation and action spaces. The output of the actor is not a Gaussian distribution as in most implementations, but a von Mises-Fisher distribution to appropriately represent the orientation of the agent (in 2D or 3D). The actor network is initialized such that initially, the output distribution is close to uniform. The main reference used in the implementation of A2C is the classical book of Sutton and Barto \cite{SuttonBarto}.

PPO also belongs to the actor-critic on-policy family of algorithms. Compared to A2C, it comes with multiple additional techniques to improve sample efficiency, learning stability, and thereby overall performance. In our implementation, which is inspired by implementations in Refs \cite{Engstrom2019,Andrychowicz2021,PPOImpl37}, we use policy loss clipping, vectorized architecture, generalized advantage estimation, advantage normalization, observation normalization, and training on fixed-length trajectory segments. The actor and critic networks are identical to those used for A2C.

For each environment and algorithm, we train ten times (ten random seeds) over $10^6$ episodes. In the TURB environment, we use 80\% of the simulation time for training, and the remaining 20\% for testing (assessing the performance of the agent in unseen flow). Hyperparameters for each algorithm were tuned manually to achieve best performance. This tuning is essential as performance is highly sensitive to some of these hyperparameters. This is the case, for example, of the learning rates but also of the parameters related to generalized advantage estimation \citep{GAE} (used in PPO). The hyperparameters we used are reported in Table \ref{tab:hyperparameters}.

\section{Results}
\label{sec:results}

\begin{figure*}
\begin{center}
    \includegraphics[width=0.99\linewidth]{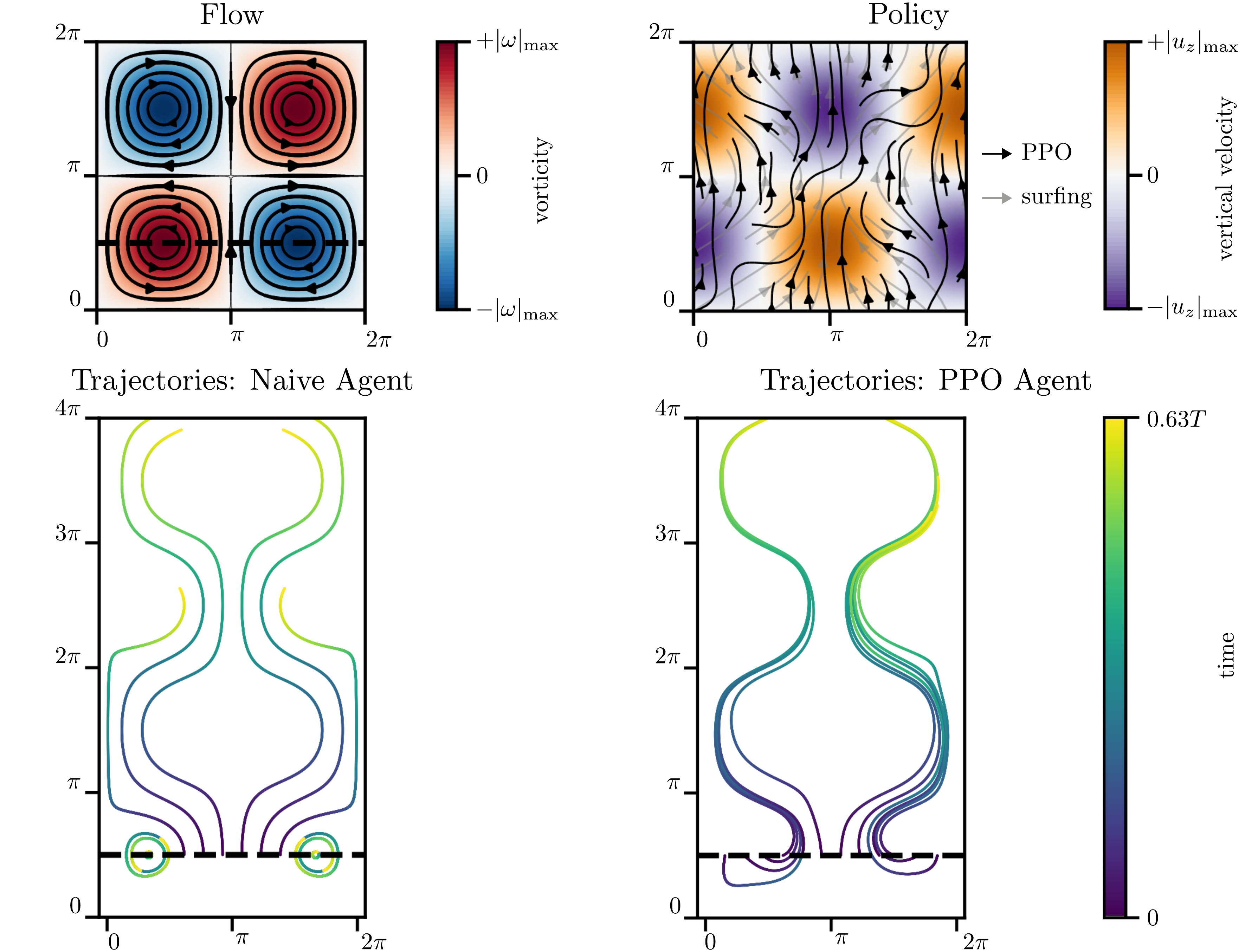}
    \caption{\label{fig:TGV}
    Navigation in Taylor-Green vortices (TGV). The flow is represented in the upper left corner by showing the (out-of-plane component of the) vorticity ($\bm{\omega} = \grad \times \bm{u}$), along with streamlines. The dashed line represents the initialization of particles whose trajectories are shown in the bottom panel, for particles following the naive strategy (left) and the learned PPO strategy (right). Unlike naive agents which can be trapped on periodic orbits, PPO agents have learned to escape such trapping and all converge to a single trajectory that yields the largest vertical displacement, independently of their initial location. The PPO policy is compared to the surfing policy in the upper right corner. While both tend to diverge from downflow regions (violet) and converge to upflow regions (orange), PPO does it more aggressively, with steeper changes of direction.}
\end{center}
\end{figure*}

\begin{figure*}
\begin{center}
    \includegraphics[width=0.99\linewidth]{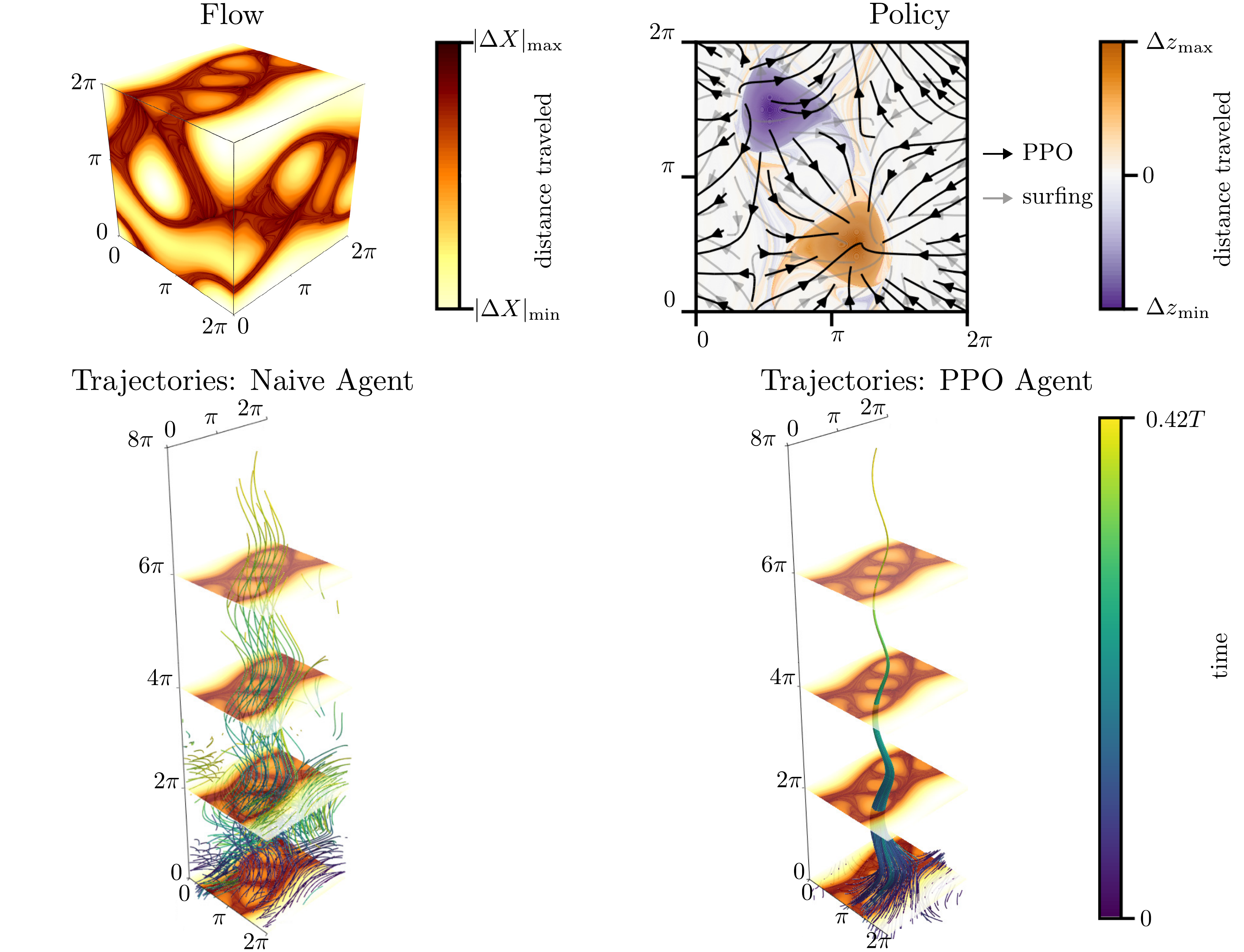}
    \caption{\label{fig:ABC}
    Navigation in Arnold-Beltrami-Childress flow (ABC). The ABC flow is represented in the upper left corner by showing the total distance travelled by passive tracers advected by the flow (image generated with \href{https://github.com/auroreloisy/ldflow}{LDflow} \revision{based on the LDDS package} \citep{LDDS}). Such quantity, called Lagrangian descriptor \citep{LagrangianDescriptors}, highlights flow regions with qualitatively different dynamics. This flow contains many coherent tube-like structure (light yellow) where tracers tend to cover large distances, these areas are also associated with preferential directions. They are separated by a chaotic region (dark red).  In the bottom panels where trajectories are shown, agents are initialized at the $z=0$ plane, following the naive strategy (left) and the PPO strategy (right). PPO agents have learned to converge to a particular flow structure, characterized by large upward transport. In the upper right corner, the surfing and PPO policies are projected onto a horizontal plane. The beneficial (detrimental) coherent structures are visible in the background: the orange (purple) one is associated to large upward (downward) displacement of passive tracers. Compared to surfing, PPO orients more aggressively toward the orange structure, which explains its overall superior performance.}
\end{center}
\end{figure*}

\begin{figure*}
\begin{center}
    \includegraphics[width=0.99\linewidth]{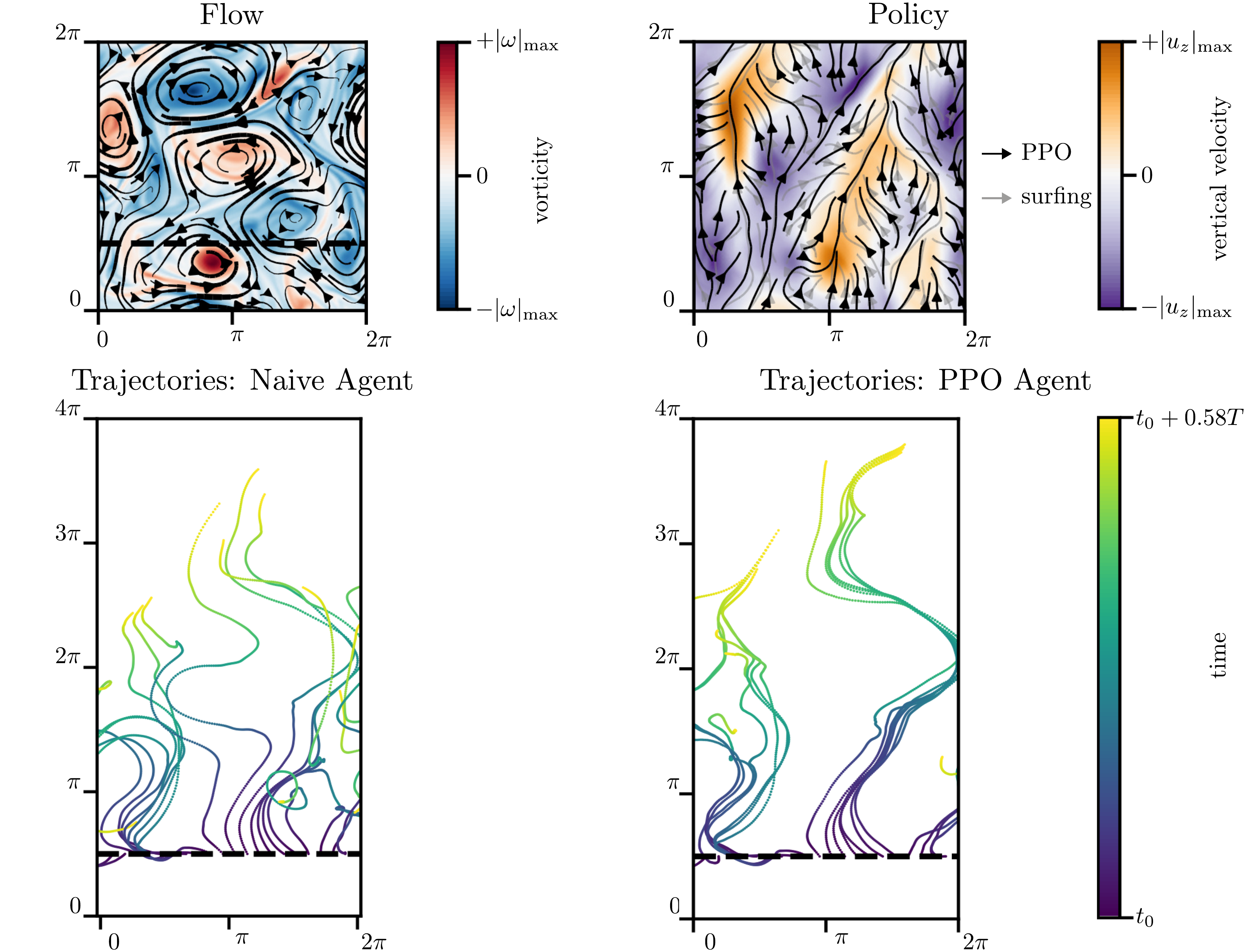}
    \caption{\label{fig:TURB}
    Navigation in a two-dimensional turbulent flow (TURB). A snapshot of the time-dependent turbulence simulation is represented in the upper left: the (out-of-plane) vorticity is shown in the background, along with the streamlines. The snapshot corresponds to a randomly chosen time $t=t_0$ at which agents are initialized on the dashed line, their trajectories are shown in the bottom panel.
    Compared to the naive strategy (bottom left), the PPO strategy (bottom right) yields trajectories that tend to clump together to benefit from upward flow. This is visible in the policy representation (top right), where PPO is compared to surfing: both tend to diverge from downflow regions (violet) and converge to upflow regions (orange). While not strictly identical, these two policies are very similar to each other.}
\end{center}
\end{figure*}

The agent's goal is to travel as far as possible in the vertical direction, by taking advantage of the partially observed carrier flow. Its performance is measured by $Z$, the vertical distance travelled over the course of an episode, averaged over all possible random initial conditions (Eq.~\ref{eq:goal}). In the following, we will present the agent performance rescaled by the naive performance: $Z/Z_{\text{naive}} = Z/(vT)$.

\begin{figure*}
\includegraphics[width=15cm]{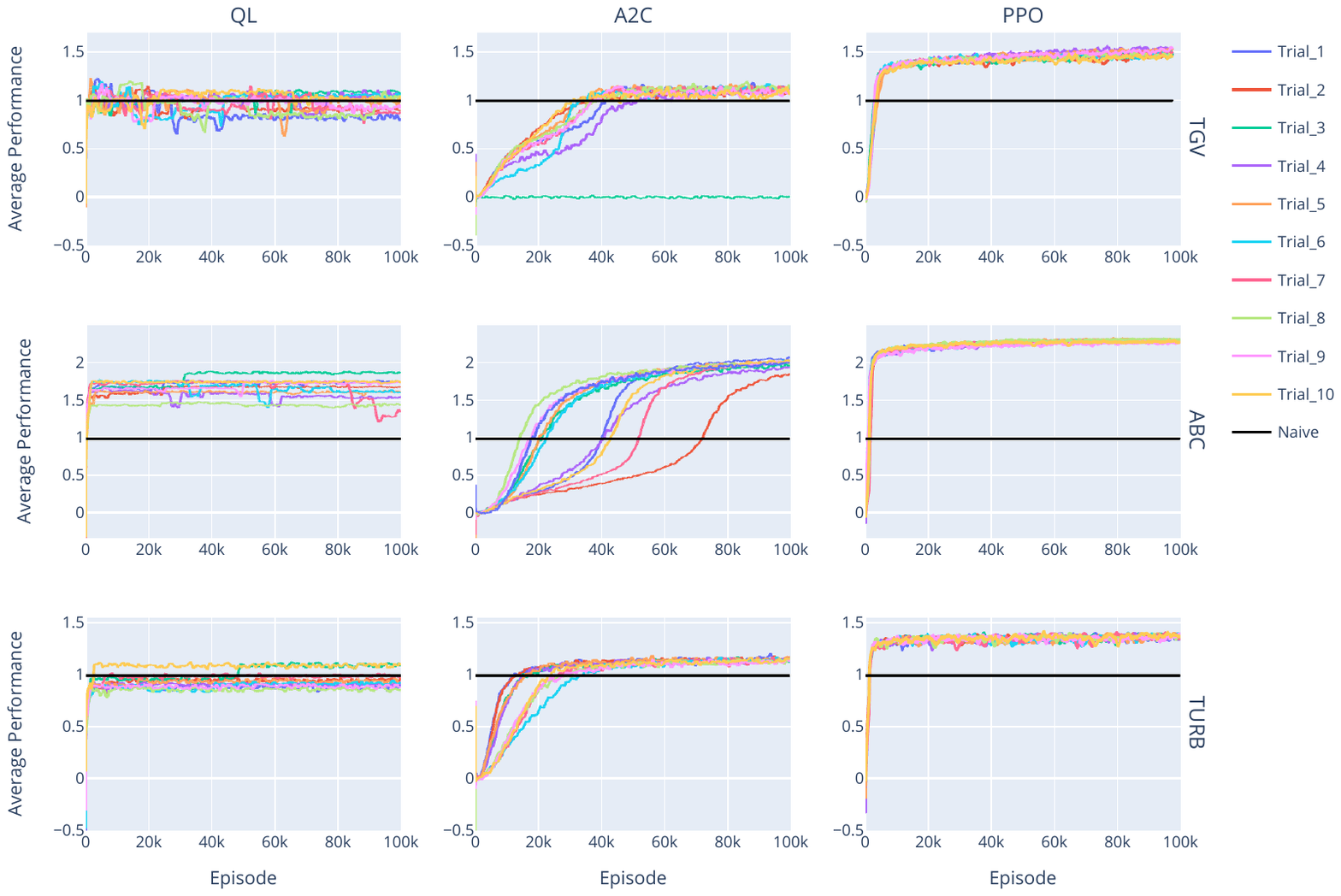}
\caption{\label{fig:learning_curves} Learning curves for each algorithm (columns) in each environment (rows): performance (distance traveled in the vertical direction normalized by $vT$) as a function of the number of episodes used for training. The performance of the naive policy is shown by a black line. To smooth out large episode-to-episode fluctuations and enhance readability, we use a moving averaging window of 1000 episodes. Only the first $10^5$ episodes are shown, training is continued over a total of $10^6$ episodes before evaluating performance in Table~\ref{tab:performance}.}
\end{figure*}

\begin{table}
\caption{Performance of the best learned policies (with 95\% confidence intervals) in the three flow environments. The performance is defined as the average vertical distance traveled in an episode, normalized by the same quantity for the naive agent.}\label{tab:performance}
\begin{tabular*}{\textwidth}{@{\extracolsep\fill}lrrrrrr}
\toprule%
  & PPO & A2C & QL & Surfing & Discrete Surfing & Naive\\ \midrule
 TGV & \textbf{1.62} $\pm$ 0.01 & 1.13 $\pm$ 0.01 & 1.22 $\pm$ 0.01 & 1.48 $\pm$ 0.01 & 1.47 $\pm$ 0.01 & 1.00 $\pm$ 0.01 \\
 ABC & \textbf{2.35} $\pm$ 0.03 & \textbf{2.32} $\pm$ 0.03 & 1.90 $\pm$ 0.03 & 2.08 $\pm$ 0.03 & 2.01 $\pm$ 0.03 & 1.00 $\pm$ 0.03 \\
 TURB & \textbf{1.51} $\pm$ 0.01 & 1.20 $\pm$ 0.01 & 1.17 $\pm$ 0.01 & \textbf{1.51} $\pm$ 0.01 & 1.39 $\pm$ 0.01 & 1.00 $\pm$ 0.01 \\
\botrule
\end{tabular*}
\end{table}

\subsection{Robustness over training trials}

Figure \ref{fig:learning_curves} shows the beginning of the learning curves (over $10^5$ episodes) of ten trials for each environment and algorithm. These learning curves allow us to assess the robustness of each algorithm. By robustness, we refer here to how repeatable are the training experiments.

Q-Learning has the largest variance across training trials. In both ABC and TURB, a single trial outperformed all the other ones. Therefore, the performance of the best agent is not easily reproducible, and strongly depends on a `lucky' random seed. In general, Q-Learning learns fast, but it is very unstable. It often unlearns good strategies as shown by sudden decreases in performance. 

A2C is found to be robust in TURB, but results were less reproducible in the other environments. While final performances are similar (with the exception of one trial in TGV where the agent did not learn anything), learning curves deviate strongly from one another. A2C tends to converge much more slowly than the other algorithms. This is due to the fact that both the actor and critic networks need to have small learning rates to ensure stability for this algorithm. The learning rates we used (cf.~Table \ref{tab:hyperparameters}) are the largest ones allowing stable convergence. 

In all the flows considered, PPO robustly reaches the same performance with very little variance across trials, compared to the other algorithms. In general, PPO converges quickly to the final solution. This is because PPO can handle high learning rates for both the actor and the critic networks, as well as training over multiple epochs because the updated policy is guaranteed not to differ too much from the original one.

\subsection{Performance of trained agents}

Table \ref{tab:performance} shows the performance of each algorithm in each environment. It is used to assess the ability of algorithms to discover good strategies in various flows. We selected the best agent for evaluation, that is, the one that achieved the highest performance at any point during training. Therefore, performance collapse during training does not affect this evaluation. In the TURB environment, we use the 20\% portion of the simulation that was not used in learning to evaluate the performance.

The evaluation was done using the deterministic version of the policies. For Q-Learning, the action chosen is the one corresponding to the highest Q-value. For A2C and PPO, instead of sampling from the von Mises-Fisher distribution that the network has converged to, we choose the action corresponding to the mean value of the distribution. Although not justified theoretically, this is common practice, and we found that the deterministic versions of the policies yield slightly better performance than the stochastic versions. Note that with PPO, the policies have essentially converged to deterministic ones, while this is not the case for A2C.

PPO is unambiguously the best-performing algorithm in all flow environments. It is the only algorithm able to outperform or match the performance of the surfing policy, our challenging baseline derived analytically, in all flows. In contrast, the strategies learned by A2C are far from optimal, except in ABC flow. As both PPO and A2C are actor-critic algorithms with identical networks, this performance gap illustrates the importance of using additional techniques (as in PPO) on top of vanilla algorithms (like A2C).

Q-Learning manages to learn better-than-naive strategies, as reported in most prior studies on microswimmer navigation. However, we are able to show here that these learned strategies are vastly suboptimal. Since Q-Learning requires discrete sets of observations and actions, we evaluated the surfing policy constrained to the same set of discrete actions (while keeping continuous observations, as discretizing observations would not make sense). Results are reported under `Discrete Surfing' in Table~\ref{tab:performance}, and show that, in all flows, Q-Learning is unable to learn a policy that matches this discrete version of the quasi-optimal strategy.


\subsection{Interpretation of the strategies learned with PPO}

We now comment on the best strategies learned for each flow, which have been obtained with the PPO algorithm.

In TGV (Fig.~\ref{fig:TGV}), naive swimmers can be trapped on periodic orbits. The occurrence of trapping, and therefore the performance of the naive agent in a given episode, is entirely determined by its initial position in the flow. In contrast, PPO agents forget their initial positions: they all converge to a single trajectory, the one that yields the largest vertical displacement by the background flow. This behaviour also prevents them from being trapped. The PPO policy and the surfing policy are similar: both diverge from downflow regions and converge to upflow regions. PPO does it more aggressively, with steeper changes of direction, resulting its slightly higher performance. We remark that, unlike surfing, the PPO policy is not symmetric with respect to symmetric inputs; this is common in reinforcement learning when no additional technique is used to enforce symmetries.

In ABC (Fig.~\ref{fig:ABC}), there exists a tube-like structure where passive tracers are trapped and are transported upward at a rapid rate (this structure is essentially an `elevator' \citep{Gustavsson2017}). Therefore, the best strategy is to get into this structure as quickly as possible from the initial position, and then get essentially carried by the upward flow. This is exactly what the PPO agent has learned to do, and its higher performance compared to other agents is directly related to its ability to reach this structure faster than all the other strategies, on average.

In TURB (Fig.~\ref{fig:TURB}), the agent needs to find and stay in regions with upward flow, without overfitting to the specific flow used for training since flow structures are random and transient. The policy learned by PPO is very similar to the analytically derived one (surfing), though not identical. \revision{To interpret this difference, we trained an agent acting according to a generalized version of the surfing policy, where the parameter $\tau^*$ is an unknown function (represented by a neural network) of the input $\bm{G}$, rather than a constant. The learned $\tau^*(\bm{G})$ varies significantly with the input values, yet the performance of this agent at the task is identical to that of the original surfing policy (and that of the PPO agent). Furthermore, we found that the policy of this generalized-surfing agent is essentially identical to that of the PPO agent. In conclusion, PPO has learned a generalized version of surfing, with a variable parameter $\tau^*$. We speculate that there is a family of functions $\tau^*(\bm{G})$ that perform as well as surfing in turbulent flows.}

\section{Conclusions}
\label{sec:conc}

We have introduced a POMDP that models a navigation task relevant to robotic microswimmers and planktonic organisms. Despite its apparent simplicity, this task is challenging because it combines complex (possibly chaotic) state dynamics with partial observability. It is nevertheless well-posed mathematically and comes with a near-optimal analytical solution. It is therefore particularly well suited as a benchmark problem for a quantitative evaluation of reinforcement learning algorithms applied to navigation in partially observable flows.

We have implemented A2C and Q-Learning with similar features as in prior studies on navigation in flows, and shown that these algorithms perform  poorly on this benchmark. In contrast, our custom implementation of PPO robustly achieves near-optimal theoretical performance. The satisfactory performance obtained with PPO is encouraging regarding the ability of reinforcement learning to discover and fully exploit flow features without having direct knowledge of them. We expect this version of PPO to be a good starting point for solving more challenging navigation tasks in partially observable flows (e.g., agents with memory).

These results highlight the importance of algorithm selection and implementation details when applying reinforcement learning to such navigation problems. We hope that these choices will be discussed more in the future, and that our study will encourage more quantitative assessments and comparisons. This could be done by comparing the performance obtained with various algorithms or by developing challenging heuristics as baselines.

The turbulent flow simulation considered here was modest to make this systematic benchmark feasible. Learning in very turbulent 2D flows and in 3D turbulent flows remains an open challenge. Analytical heuristics such as the surfing policy \citep{Monthiller2022} or its recent generalizations \citep{Calascibetta2023b,Piro2024} provide strong baselines to which learned strategies should be compared to. As the cost of running the environment increases, PPO may become inefficient. Off-policy algorithms, such as SAC (Soft Actor Critic \citep{SAC}) and TD3 (Twin-Delayed Deep Deterministic policy gradient \citep{TD3}) should be considered and benchmarked for such problems where sample efficiency is likely to be of critical importance.

Partial observability is sometimes counteracted by providing the agent with some form of memory. While memory is unnecessary for the navigation task considered here, it is crucial to other navigation problems such as olfactory search in turbulent flows \cite{Vergassola2007,Loisy2022a,Loisy2023,Reddy2022b,Celani2024}, a much harder problem on which reinforcement learning has started to be used \citep{Verano2023,Singh2023,Rando2025preprint}. It remains to be shown whether model-free, deep reinforcement learning is a viable tool for discovering good strategies in such memory-based navigation tasks in the presence of a realistic turbulent flow.

\backmatter

\bmhead{Acknowledgments}
We thank Rémi Monthiller for fruitful discussions on optimal planktonic navigation, Jérémie Bec for his help with early developments of the DNS code for turbulent flow simulations, and Vladimír Krajňák for his help with visualizing coherent structures in ABC flow using Lagrangian descriptors.
This project has received funding from the European Research Council (ERC) under the European Union's Horizon 2020 research and innovation programme (grant agreement No 834238). 

\section*{Statements and Declarations}

\subsection*{Competing interests}
The authors have no competing interests to declare.

\subsection*{Code and data availability}
The code used for this study and the trained models are available at \url{https://github.com/C0PEP0D/RLfl0w}.

\subsection*{Authors' contributions}
SM, AL and CE designed the study, analyzed the results and edited the manuscript. SM developed the code, performed numerical experiments and processed results. AL drafted the manuscript. CE obtained funding.

\newpage
\begin{appendices}

\section{Parameters used for the surfing policy, the actor-critic neural networks, and the reinforcement learning algorithms}

\begin{figure*}[h]
\begin{center}
\includegraphics[width=13cm]{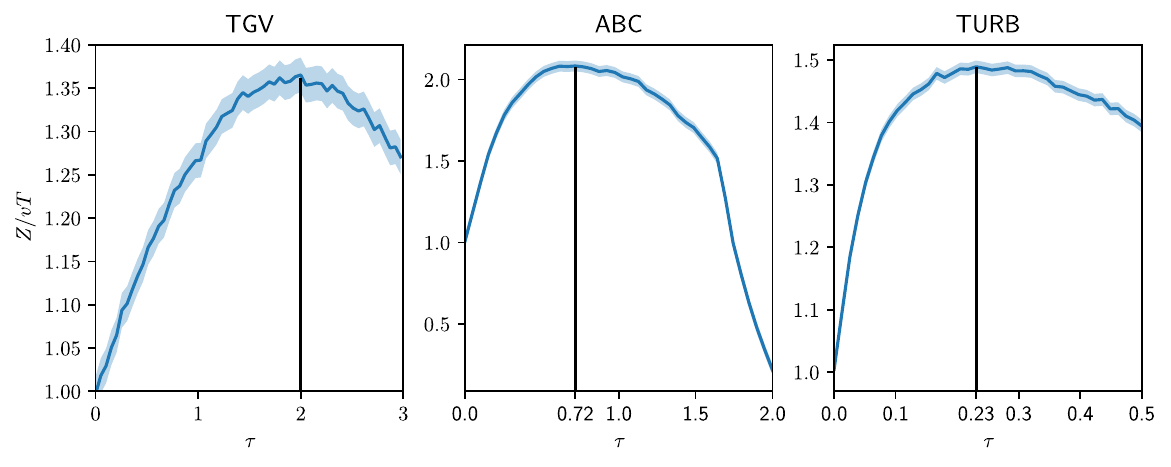}
\caption{\label{fig:surf_opt}Surfing policy: optimization over $\tau$ in the different flows. In TGV $\tau^{*} = 2.0$, in ABC $\tau^{*} = 0.72$, and in TURB $\tau^{*} = 0.23$}
\end{center}
\end{figure*}

\begin{table}[h]
\caption{\label{tab:nn}Parameters of the actor and critic neural networks, used in both A2C and PPO. The two networks are independent (no shared layer).}
\begin{tabular}{p{5cm}p{3cm}}
\toprule
 Actor network & \\ \midrule
 Number of hidden layers & 2\\
 Neurons per hidden layer & 40\\
 Type of layers & Dense\\
 Initialization & Glorot-uniform\\
 Activation & ELU\\
 Use feature normalization & True\\
 Optimizer & Adam\\
 Output distribution & von Mises-Fisher\\
 \botrule
\end{tabular}
\begin{tabular}{p{5cm}p{3cm}}
\toprule
 Critic network & \\ \midrule
 Number of hidden layers & 2\\
 Neurons per hidden layer & 100\\
 Type of layers & Dense\\
 Initialization & Glorot-uniform\\
 Activation & ELU\\
 Use feature normalization & True\\
 Optimizer & Adam\\
 \botrule
\end{tabular}
\end{table}

\begin{table}[h]
\caption{\label{tab:hyperparameters}Hyperparameters used for the learning algorithms.}
\begin{tabular}{p{5cm}p{1cm}p{1cm}p{1cm}}
\toprule
Hyperparameter of QL & TGV & ABC & TURB\\ \midrule
Learning rate & 0.8 & 0.8 & 0.8\\
Anneal learning rates & True & True & True\\
Epsilon (for $\epsilon$-greedy exploration) & 0.1 & 0.1 & 0.1\\
Discount factor & 0.95 & 0.95 & 0.99\\
\botrule
\end{tabular}
\begin{tabular}{p{5cm}p{1cm}p{1cm}p{1cm}}
\toprule
Hyperparameter of A2C & TGV & ABC & TURB\\ \midrule
Learning rate actor & $10^{-6}$ & $10^{-6}$ & $10^{-6}$\\
Learning rate critic & $10^{-4}$ & $10^{-4}$ & $10^{-4}$\\
Anneal learning rates & False & False & False\\
Discount factor & 0.95 & 0.99 & 0.99\\
\botrule
\end{tabular}
\begin{tabular}{p{5cm}p{1cm}p{1cm}p{1cm}}
\toprule
Hyperparameter of PPO & TGV & ABC & TURB\\ \midrule
Learning rate actor & $10^{-4}$ & $10^{-4}$ & $10^{-4}$\\
Learning rate critic & $10^{-3}$ & $10^{-3}$ & $10^{-3}$\\
Anneal learning rates & True & True & True\\
Discount factor & 0.99 & 0.99 & 0.99\\
Number of parallel environments & 100 & 10 & 10\\
Rollout length & 10 & 10 & 100\\
GAE lambda & 1.0 & 1.0 & 1.0\\
Number of minibatches & 5 & 5 & 5\\
Epochs & 4 & 4 & 4\\
Clip coefficient & 0.1 & 0.1 & 0.1\\
Entropy coefficient & 0.0 & 0.0 & 0.0\\
Target KL divergence & 0.02 & 0.02 & 0.02\\
\botrule
\end{tabular}
\end{table}
\end{appendices}

\newpage

\newpage
\bibliography{sn-bibliography}

\begin{thebibliography}{10}
\providecommand{\url}[1]{{#1}}
\providecommand{\urlprefix}{URL }
\providecommand{\doi}[1]{\url{https://doi.org/#1}}
\bibcommenthead

\bibitem{MS1}
{B. Dai}, {J. Wang}, {Z. Xiong}, {X. Zhan}, {W. Dai}, {C. Li}, {S. Feng}, {J.
  Tang}, Programmable artificial phototactic microswimmer.
\newblock Nature Nanotechnology \textbf{11}, 1087--1092 (2016)

\bibitem{MS2}
{H. Huang}, {F. Uslu}, {P. Katsamba}, {E. Lauga}, {M. Sakar}, {B. Nelson},
  Adaptive locomotion of artificial microswimmers.
\newblock Science Advances \textbf{5} (2019)

\bibitem{Muinos-Landin2021}
S.~{Mui{\~n}os-Landin}, A.~Fischer, V.~Holubec, F.~Cichos, Reinforcement
  learning with artificial microswimmers.
\newblock Science Robotics \textbf{6}(52) (2021).
\newblock \doi{10.1126/scirobotics.abd9285}

\bibitem{BioGrads1}
{J. Wheeler}, {E. Secchi}, {R. Rusconi}, {R. Stocker}, Not just going with the
  flow: The effects of fluid flow on bacteria and plankton.
\newblock Annual Review of Cell and Developmental Biology \textbf{35}, 213--237
  (2019)

\bibitem{BioGrads2}
{T. Ki{\o}rboe}, {E. Saiz}, {A. Visser}, Hydrodynamic signal perception in the
  copepod {A}cartia tonsa.
\newblock Marine Ecology Progress Series \textbf{179}, 97--111 (1999)

\bibitem{Monthiller2022}
R.~Monthiller, A.~Loisy, M.A.R. Koehl, B.~Favier, C.~Eloy, Surfing on
  {{Turbulence}}: {{A Strategy}} for {{Planktonic Navigation}}.
\newblock Physical Review Letters \textbf{129}(6), 064502 (2022).
\newblock \doi{10.1103/PhysRevLett.129.064502}

\bibitem{Zermelo}
{E. Zermelo}, {Ü}ber das {N}avigationsproblem bei ruhender oder
  veränderlicher {W}indverteilung.
\newblock Journal of Applied Mathematics and Mechanics \textbf{11}, 114--124
  (1931)

\bibitem{Gazzola2014}
M.~Gazzola, B.~Hejazialhosseini, P.~Koumoutsakos, Reinforcement {{Learning}}
  and {{Wavelet Adapted Vortex Methods}} for {{Simulations}} of
  {{Self-propelled Swimmers}}.
\newblock SIAM Journal on Scientific Computing \textbf{36}(3), B622--B639
  (2014).
\newblock \doi{10.1137/130943078}

\bibitem{Reddy2016a}
G.~Reddy, A.~Celani, T.J. Sejnowski, M.~Vergassola, Learning to soar in
  turbulent environments.
\newblock Proceedings of the National Academy of Sciences of the United States
  of America \textbf{113}(33), E4877--84 (2016).
\newblock \doi{10.1073/pnas.1606075113}.
\newblock {\href{https://arxiv.org/abs/27482099}{{27482099}}}

\bibitem{Gazzola2016}
M.~Gazzola, A.A. Tchieu, D.~Alexeev, A.~de~Brauer, P.~Koumoutsakos, Learning to
  school in the presence of hydrodynamic interactions.
\newblock Journal of Fluid Mechanics \textbf{789}, 726--749 (2016).
\newblock \doi{10.1017/jfm.2015.686}

\bibitem{Colabrese2017}
S.~Colabrese, K.~Gustavsson, A.~Celani, L.~Biferale, Flow {{Navigation}} by
  {{Smart Microswimmers}} via {{Reinforcement Learning}}.
\newblock Physical Review Letters \textbf{118}(15), 158004 (2017).
\newblock \doi{10.1103/PhysRevLett.118.158004}

\bibitem{Gustavsson2017}
K.~Gustavsson, L.~Biferale, A.~Celani, S.~Colabrese, Finding efficient swimming
  strategies in a three-dimensional chaotic flow by reinforcement learning.
\newblock The European Physical Journal E \textbf{40}(12), 110--110 (2017).
\newblock \doi{10.1140/epje/i2017-11602-9}

\bibitem{Reddy2018}
G.~Reddy, J.~{Wong-Ng}, A.~Celani, T.J. Sejnowski, M.~Vergassola, Glider
  soaring via reinforcement learning in the field.
\newblock Nature \textbf{562}(7726), 236--239 (2018).
\newblock \doi{10.1038/s41586-018-0533-0}

\bibitem{Colabrese2018}
S.~Colabrese, K.~Gustavsson, A.~Celani, L.~Biferale, Smart inertial particles.
\newblock Physical Review Fluids \textbf{3}(8), 084301 (2018).
\newblock \doi{10.1103/PhysRevFluids.3.084301}

\bibitem{Verma2018}
S.~Verma, G.~Novati, P.~Koumoutsakos, Efficient collective swimming by
  harnessing vortices through deep reinforcement learning.
\newblock Proceedings of the National Academy of Sciences \textbf{115}(23),
  5849--5854 (2018).
\newblock \doi{10.1073/pnas.1800923115}

\bibitem{Alageshan2020}
J.K. Alageshan, A.K. Verma, J.~Bec, R.~Pandit, Machine learning strategies for
  path-planning microswimmers in turbulent flows.
\newblock Physical Review E \textbf{101}(4), 043110--043110 (2020).
\newblock \doi{10.1103/PhysRevE.101.043110}

\bibitem{Gunnarson2021}
P.~Gunnarson, I.~Mandralis, G.~Novati, P.~Koumoutsakos, J.O. Dabiri, Learning
  efficient navigation in vortical flow fields.
\newblock Nature Communications \textbf{12}(1), 7143 (2021).
\newblock \doi{10.1038/s41467-021-27015-y}

\bibitem{Mandralis2021}
I.~Mandralis, P.~Weber, G.~Novati, P.~Koumoutsakos, Learning swimming escape
  patterns for larval fish under energy constraints.
\newblock Physical Review Fluids \textbf{6}(9), 093101 (2021).
\newblock \doi{10.1103/PhysRevFluids.6.093101}

\bibitem{Qiu2022a}
J.~Qiu, N.~Mousavi, K.~Gustavsson, C.~Xu, B.~Mehlig, L.~Zhao, Navigation of
  micro-swimmers in steady flow: The importance of symmetries.
\newblock Journal of Fluid Mechanics \textbf{932} (2022).
\newblock \doi{10.1017/jfm.2021.978}

\bibitem{Qiu2022b}
J.~Qiu, N.~Mousavi, L.~Zhao, K.~Gustavsson, Active gyrotactic stability of
  microswimmers using hydromechanical signals.
\newblock Physical Review Fluids \textbf{7}(1), 014311 (2022).
\newblock \doi{10.1103/PhysRevFluids.7.014311}

\bibitem{Calascibetta2023a}
C.~Calascibetta, L.~Biferale, F.~Borra, A.~Celani, M.~Cencini, Taming
  {{Lagrangian}} chaos with multi-objective reinforcement learning.
\newblock The European Physical Journal E \textbf{46}(3), 9 (2023).
\newblock \doi{10.1140/epje/s10189-023-00271-0}

\bibitem{ElKhiyati2023}
Z.~El~Khiyati, R.~Chesneaux, L.~Giraldi, J.~Bec, Steering undulatory
  micro-swimmers in a fluid flow through reinforcement learning.
\newblock The European Physical Journal E \textbf{46}(6), 43 (2023).
\newblock \doi{10.1140/epje/s10189-023-00293-8}

\bibitem{Sankaewtong2023}
K.~Sankaewtong, J.J. Molina, M.S. Turner, R.~Yamamoto, Learning to swim
  efficiently in a nonuniform flow field.
\newblock Physical Review E \textbf{107}(6), 065102 (2023).
\newblock \doi{10.1103/PhysRevE.107.065102}

\bibitem{Gunnarson2024}
P.~Gunnarson, J.O. Dabiri, Fish-inspired tracking of underwater turbulent
  plumes.
\newblock Bioinspiration \& Biomimetics \textbf{19}(5), 056024 (2024).
\newblock \doi{10.1088/1748-3190/ad7181}

\bibitem{Mousavi2025}
N.~Mousavi, J.~Qiu, L.~Zhao, B.~Mehlig, K.~Gustavsson, Short term vs. long
  term: {{Optimization}} of microswimmer navigation on different time horizons.
\newblock Physical Review Research \textbf{7}(1), 013258 (2025).
\newblock \doi{10.1103/PhysRevResearch.7.013258}

\bibitem{Jiao2025}
Y.~Jiao, H.~Hang, J.~Merel, E.~Kanso, Sensing flow gradients is necessary for
  learning autonomous underwater navigation.
\newblock Nature Communications \textbf{16}(1), 3044 (2025).
\newblock \doi{10.1038/s41467-025-58125-6}

\bibitem{QL}
C.J.C.H. Watkins, Learning from delayed rewards.
\newblock Ph.D. thesis, University of Cambridge, England (1989)

\bibitem{A2C}
V.~Mnih, A.P. Badia, M.~Mirza, A.~Graves, T.~Lillicrap, T.~Harley, D.~Silver,
  K.~Kavukcuoglu, \emph{Asynchronous {{Methods}} for {{Deep Reinforcement
  Learning}}}, in \emph{Proceedings of {{The}} 33rd {{International
  Conference}} on {{Machine Learning}}} (PMLR, 2016), pp. 1928--1937

\bibitem{PPO}
J.~Schulman, F.~Wolski, P.~Dhariwal, A.~Radford, O.~Klimov.
\newblock Proximal {{Policy Optimization Algorithms}} (2017).
\newblock \doi{10.48550/arXiv.1707.06347}

\bibitem{Boffetta2012}
G.~Boffetta, R.E. Ecke, Two-{{Dimensional Turbulence}}.
\newblock Annual Review of Fluid Mechanics \textbf{44}(1), 427--451 (2012).
\newblock \doi{10.1146/annurev-fluid-120710-101240}

\bibitem{Biferale2019}
L.~Biferale, F.~Bonaccorso, M.~Buzzicotti, P.~Clark Di~Leoni, K.~Gustavsson,
  Zermelo's problem: {{Optimal}} point-to-point navigation in {{2D}} turbulent
  flows using reinforcement learning.
\newblock Chaos: An Interdisciplinary Journal of Nonlinear Science
  \textbf{29}(10), 103138--103138 (2019).
\newblock \doi{10.1063/1.5120370}

\bibitem{SuttonBarto}
R.S. Sutton, A.G. Barto, \emph{Reinforcement {{Learning}}: {{An
  Introduction}}}, 2nd edn. (MIT Press, Cambridge, MA, 2018)

\bibitem{Engstrom2019}
L.~Engstrom, A.~Ilyas, S.~Santurkar, D.~Tsipras, F.~Janoos, L.~Rudolph,
  A.~Madry, \emph{Implementation {{Matters}} in {{Deep RL}}: {{A Case Study}}
  on {{PPO}} and {{TRPO}}}, in \emph{{{ICLR}} 2019 - {{Eighth International
  Conference}} on {{Learning Representations}}} (2019)

\bibitem{Andrychowicz2021}
M.~Andrychowicz, A.~Raichuk, P.~Sta{\'n}czyk, M.~Orsini, S.~Girgin,
  R.~Marinier, L.~Hussenot, M.~Geist, O.~Pietquin, M.~Michalski, S.~Gelly,
  O.~Bachem, \emph{What {{Matters In On-Policy Reinforcement Learning}}? {{A
  Large-Scale Empirical Study}}}, in \emph{{{ICLR}} 2021 - {{Ninth
  International Conference}} on {{Learning Representations}}} (Vienna, Austria,
  2021)

\bibitem{PPOImpl37}
{H. Shengyi}, {D. Julien}, {R. Atonin}, {K. Anssi}, W..I.B. Track.
\newblock The 37 implementation details of proximal policy optimization (2022).
\newblock
  \urlprefix\url{https://iclr-blog-track.github.io/2022/03/25/ppo-implementation-details/}

\bibitem{GAE}
J.~Schulman, P.~Moritz, S.~Levine, M.~Jordan, P.~Abbeel,
  \emph{High-{{Dimensional Continuous Control Using Generalized Advantage
  Estimation}}}, in \emph{Proceedings of the 4th {{International Conference}}
  on {{Learning Representations}} ({{ICLR}} 2016)} (arXiv, San Juan, Puerto
  Rico, 2016).
\newblock \doi{10.48550/arXiv.1506.02438}

\bibitem{LDDS}
B.~Aguilar-Sanjuan, V.J. García-Garrido, V.~Krajňák, S.~Naik, S.~Wiggins,
  {LDDS}: Python package for computing and visualizing lagrangian descriptors
  for dynamical systems.
\newblock Journal of Open Source Software \textbf{6}(65), 3482 (2021).
\newblock \doi{10.21105/joss.03482}.
\newblock \urlprefix\url{https://doi.org/10.21105/joss.03482}

\bibitem{LagrangianDescriptors}
M.~Agaoglou, B.~Aguilar~Sanjuan, V.J. Garc{\'i}a-Garrido, F.~{Gonzalez
  Montoya}, M.~Katsanikas, V.~Kraj{\v n}{\'a}k, S.~Naik, S.R. Wiggins,
  \emph{{Lagrangian Descriptors: Discovery and Quantification of Phase Space
  Structure and Transport}} (Zenodo, 2020).
\newblock \doi{10.5281/zenodo.3958985}

\bibitem{Calascibetta2023b}
C.~Calascibetta, L.~Biferale, F.~Borra, A.~Celani, M.~Cencini, Optimal tracking
  strategies in a turbulent flow.
\newblock Communications Physics \textbf{6}(1), 1--10 (2023).
\newblock \doi{10.1038/s42005-023-01366-y}

\bibitem{Piro2024}
L.~Piro, A.~Vilfan, R.~Golestanian, B.~Mahault, Energetic cost of microswimmer
  navigation: {{The}} role of body shape.
\newblock Physical Review Research \textbf{6}(1), 013274 (2024).
\newblock \doi{10.1103/PhysRevResearch.6.013274}

\bibitem{SAC}
T.~Haarnoja, A.~Zhou, P.~Abbeel, S.~Levine, \emph{Soft {{Actor-Critic}}:
  {{Off-Policy Maximum Entropy Deep Reinforcement Learning}} with a
  {{Stochastic Actor}}}, in \emph{Proceedings of the 35th {{International
  Conference}} on {{Machine Learning}}} (PMLR, Stockholm, Sweden, 2018), pp.
  1861--1870

\bibitem{TD3}
S.~Fujimoto, H.~Hoof, D.~Meger, \emph{Addressing {{Function Approximation
  Error}} in {{Actor-Critic Methods}}}, in \emph{Proceedings of the 35th
  {{International Conference}} on {{Machine Learning}}} (PMLR, Stockholm,
  Sweden, 2018), pp. 1587--1596

\bibitem{Vergassola2007}
M.~Vergassola, E.~Villermaux, B.I. Shraiman, "{{Infotaxis}}" as a strategy for
  searching without gradients.
\newblock Nature \textbf{445}(7126), 406--409 (2007).
\newblock \doi{10.1038/nature05464}

\bibitem{Loisy2022a}
A.~Loisy, C.~Eloy, Searching for a source without gradients: How good is
  infotaxis and how to beat it.
\newblock Proceedings of the Royal Society A: Mathematical, Physical and
  Engineering Sciences \textbf{478}(2262), 20220118 (2022).
\newblock \doi{10.1098/rspa.2022.0118}

\bibitem{Loisy2023}
A.~Loisy, R.A. Heinonen, Deep reinforcement learning for the olfactory search
  {{POMDP}}: A quantitative benchmark.
\newblock The European Physical Journal E \textbf{46}(3), 17 (2023).
\newblock \doi{10.1140/epje/s10189-023-00277-8}

\bibitem{Reddy2022b}
G.~Reddy, V.N. Murthy, M.~Vergassola, Olfactory {{Sensing}} and {{Navigation}}
  in {{Turbulent Environments}}.
\newblock Annual Review of Condensed Matter Physics \textbf{13}(1), 191--213
  (2022).
\newblock \doi{10.1146/annurev-conmatphys-031720-032754}

\bibitem{Celani2024}
A.~Celani, E.~Panizon, in \emph{Target {{Search Problems}}}, ed. by
  D.~Grebenkov, R.~Metzler, G.~Oshanin (Springer Nature Switzerland, 2024), pp.
  711--732.
\newblock \doi{10.1007/978-3-031-67802-8_30}

\bibitem{Verano2023}
K.V.B. Verano, E.~Panizon, A.~Celani, Olfactory search with finite-state
  controllers.
\newblock Proceedings of the National Academy of Sciences \textbf{120}(34),
  e2304230120 (2023).
\newblock \doi{10.1073/pnas.2304230120}

\bibitem{Singh2023}
S.H. Singh, F.~{van Breugel}, R.P.N. Rao, B.W. Brunton, Emergent behaviour and
  neural dynamics in artificial agents tracking odour plumes.
\newblock Nature Machine Intelligence \textbf{5}(1), 58--70 (2023).
\newblock \doi{10.1038/s42256-022-00599-w}

\bibitem{Rando2025preprint}
M.~Rando, M.~James, A.~Verri, L.~Rosasco, A.~Seminara, Q-learning with temporal
  memory to navigate turbulence.
\newblock eLife \textbf{13} (2025).
\newblock \doi{10.7554/eLife.102906.2}

\end{thebibliography}

\end{document}